\PassOptionsToPackage{table}{xcolor}
\documentclass[11pt]{article}

\usepackage{acl}

\usepackage{times}
\usepackage{latexsym}
\usepackage[T1]{fontenc}
\usepackage[utf8]{inputenc}
\usepackage{microtype}
\usepackage{inconsolata}
\usepackage{graphicx}

\usepackage{amsmath}
\usepackage{amssymb}
\usepackage{booktabs}
\usepackage{float}
\usepackage{multirow}
\usepackage{array}
\usepackage{tabularx}
\usepackage{adjustbox}
\usepackage{makecell}
\usepackage{siunitx}
\usepackage{xspace}

\title{IntentVLA: Short-Horizon Intent Modeling for Aliased Robot Manipulation}

\author{
  Shijie Lian\textsuperscript{1,2,10\thanks{Equal contribution}}
  Bin Yu\textsuperscript{2,4,10\footnotemark[1]}
  Xiaopeng Lin\textsuperscript{5,10\footnotemark[1]}
  Zhaolong Shen\textsuperscript{2,6,10\footnotemark[1]}
  Laurence Tianruo Yang\textsuperscript{1,7,\thanks{Corresponding authors}}\\
  \textbf{Yurun Jin}\textsuperscript{9}
  \textbf{Haishan Liu}\textsuperscript{2}
  \textbf{Changti Wu\textsuperscript{2,8}}
  \textbf{Hang Yuan\textsuperscript{2,8}}
  \textbf{Cong Huang\textsuperscript{2,3}}
  \textbf{Kai Chen\textsuperscript{2,3,10,\footnotemark[2]}}
  \\[2ex]
  \textsuperscript{1}Huazhong University of Science and Technology\\
  \textsuperscript{2}Zhongguancun Academy\quad
  \textsuperscript{3}Zhongguancun Institute of Artificial Intelligence\\
  \textsuperscript{4}Harbin Institute of Technology\quad
  \textsuperscript{5}The Hong Kong University of Science and Technology (Guangzhou)\\
  \textsuperscript{6}Beihang University\quad
  \textsuperscript{7}Zhengzhou University\quad
  \textsuperscript{8}East China Normal University\\
  \textsuperscript{9}University of Science and Technology of China\quad
  \textsuperscript{10}DeepCybo
}

\begin{document}
\maketitle

\let\thefootnote\relax\footnotetext{Work done at Zhongguancun Academy (Beijing).}

\newcommand{\methodname}{IntentVLA\xspace}
\newcommand{\benchmarkname}{AliasBench\xspace}
\newcommand{\taskcell}[2]{\makecell[l]{\texttt{#1}\\\texttt{#2}}}
\newcommand{\gain}[1]{\,\textcolor{green!60!black}{$\uparrow$\scriptsize{#1}}}
\newcommand{\gainph}[1]{\phantom{\,\textcolor{green!60!black}{$\uparrow$\scriptsize{#1}}}}

\begin{abstract}
Robot imitation data are often multimodal: similar visual-language observations may be followed by different action chunks because human demonstrators act with different short-horizon intents, task phases, or recent context. Existing frame-conditioned VLA policies infer each chunk from the current observation and instruction alone, so under partial observability they may resample different intents across adjacent replanning steps, leading to inter-chunk conflict and unstable execution. We introduce \methodname, a history-conditioned VLA framework that encodes recent visual observations into a compact short-horizon intent representation and uses it to condition chunk generation. We further introduce \benchmarkname, a 12-task ambiguity-aware benchmark on RoboTwin2 with matched training data and evaluation environments that isolate short-horizon observation aliasing. Across \benchmarkname, SimplerEnv, LIBERO, and RoboCasa, \methodname{} improves rollout stability and outperforms strong VLA baselines. The benchmark code, generated data, and model code are released at \url{https://github.com/ZGC-EmbodyAI/IntentVLA}.
\end{abstract}

\section{Introduction}
\label{sec:intro}

Vision-language-action (VLA) models provide a direct interface from perception and instruction to control: given visual observations and a language command, the policy outputs robot actions~\cite{OpenVLA_2024_CoRL,PI0_2024_arXiv,RDT1B_2025_ICLR,GR00T_2025_arXiv}.
Recent large-model-based VLAs scale this paradigm with transformer backbones, large robot datasets, and vision-language pretraining, enabling more generalist manipulation policies across tasks and embodiments \cite{PI05_2025_arXiv,GR00T_N1.6,HRDT_2025_arXiv,X-VLA_2025_arXiv,RDT2_2026_arXiv}.

Training VLA models typically relies on large-scale human-collected robot trajectories \cite{OXE_2024_ICRA,Agibot_2025_arXiv,Bridgedatav2_2023_CoRL,Droid_2024_arXiv}, and these datasets often faithfully reflect the underlying multimodality of manipulation behavior. 
For instance, an environment may admit multiple valid goals, and even a fixed goal can often be achieved through multiple feasible paths \cite{zhai2025vfp}. 
This diversity is not itself the problem. Human demonstrations are naturally multimodal across episodes, but they are locally committed within each episode: once a demonstrator follows a particular task phase, path, or completion strategy, adjacent action chunks usually remain consistent with that choice. The difficulty arises because current VLA policies generally infer actions from only the current frame image and the language instruction. Under partial observability, the same frame-level observation can correspond to different short-horizon intents, but a frame-conditioned VLA does not observe the episode-level commitment that selected one of them. Repeated chunk generation can then switch among intents across adjacent decision steps, producing contradictory chunks and unstable execution. Thus, the goal is not to eliminate multimodality, but to condition generation on the commitment already expressed by the current episode.

\begin{figure*}[t]
    \centering
    \includegraphics[width=1.0\textwidth]{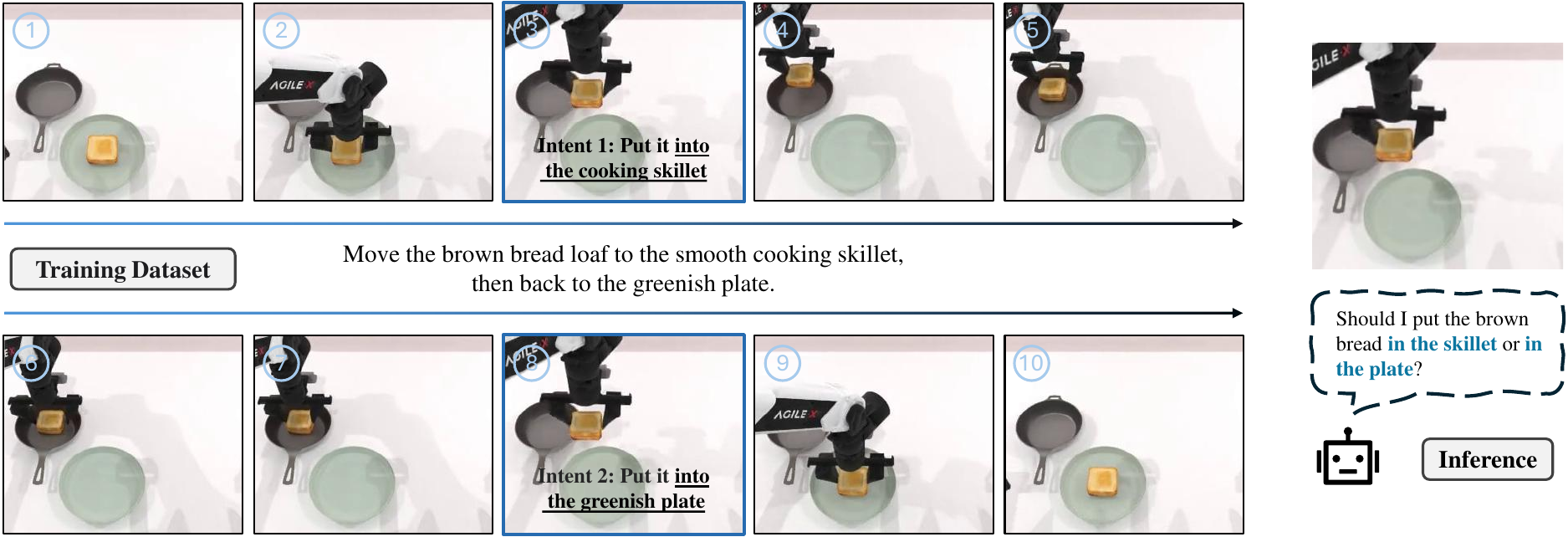}
    \caption{\textbf{Short-horizon intent ambiguity under frame-only conditioning.} Similar bread-in-gripper observations require different continuations: cooking in the skillet or returning to the plate.}
    \label{fig:multimodality_show1}
\end{figure*}

\begin{figure*}[t]
    \centering
    \includegraphics[width=1.0\textwidth]{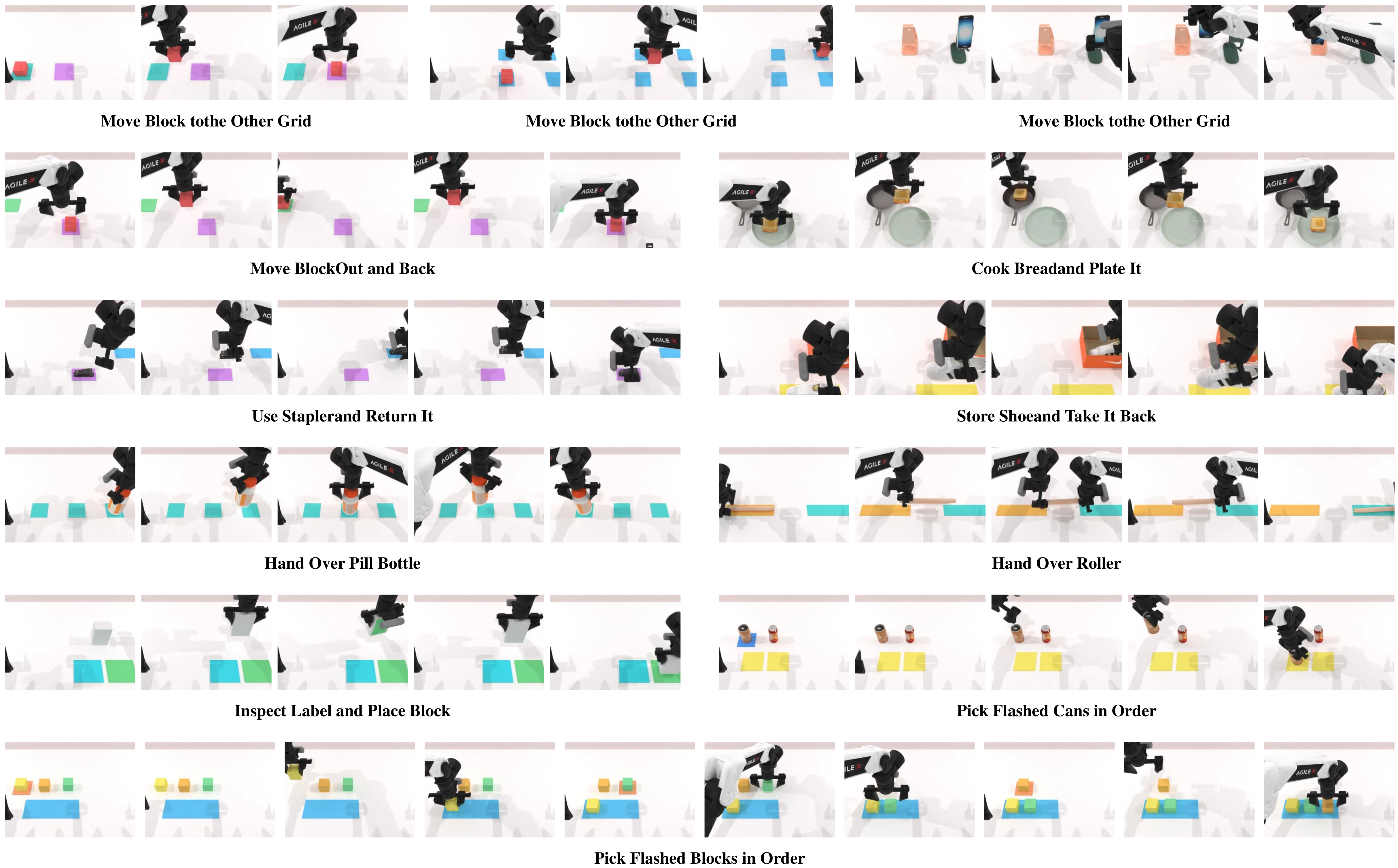}
    \caption{\textbf{Representative observation aliasing patterns in \benchmarkname{}.} The quantitative observation-aliasing diagnostic is shown in Figure~\ref{fig:aliasbench_aliasing_diagnostic}, and policy results are reported in Section~\ref{sec:aliasbench_results}.}
    \label{fig:aliasbench_cases}
\end{figure*}

Figure~\ref{fig:multimodality_show1} illustrates this ambiguity: similar bread-holding observations can require skillet placement or plate return under the same instruction. To measure this failure mode, we build \benchmarkname{} on RoboTwin2 \cite{RoboTwin2_2025_arXiv} with matched simulation training data and evaluation environments that isolate short-horizon observation aliasing. Rather than only scoring task success, \benchmarkname{} tests whether policies preserve local continuations when visually similar states require different next chunks. It covers back-and-forth, crossing-path, bimanual, and multi-goal ambiguity, with examples in Figure~\ref{fig:aliasbench_cases}.

To address the failure mode validated by \benchmarkname{}, we propose \methodname, a history-conditioned imitation learning framework for chunked VLA control. The core idea is to preserve the episode's local commitment by conditioning action generation on recent visual evidence, rather than inferring every chunk from the current frame alone. Concretely, \methodname{} encodes recent observations with a frozen VGGT-based history encoder, keeps compact camera and register tokens as history evidence, and fuses them with the current Qwen3-VL visual-language context through gated cross-attention. The fused context and an appended history-evidence token form a short-horizon intent representation that conditions a standard DiT-based flow-matching action head. In experiments on \benchmarkname{}, SimplerEnv, LIBERO, and RoboCasa, \methodname{} improves both success rate and execution stability over strong VLA baselines.

Our contributions are fourfold:
\begin{itemize}
    \item We identify a failure mode of frame-conditioned chunk policies under partial observability: demonstrations are multimodal across episodes but locally committed within an episode, while frame-only conditioning can break this commitment at test time.
    \item We construct \benchmarkname, a 12-task benchmark on RoboTwin2 for evaluating VLA behavior under short-horizon observation aliasing, together with matched simulation training data and evaluation environments.
    \item We propose \methodname, a history-conditioned imitation learning framework that learns a compact short-horizon intent representation from recent visual observations and uses it to condition chunk generation.
    \item We implement and validate \methodname{} extensively across \benchmarkname{}, SimplerEnv, LIBERO, and RoboCasa, including ambiguous-intent tasks that directly test short-horizon intent consistency.
\end{itemize}

\section{Related Work}
\label{sec:related}
\subsection{Vision-Language-Action Models}
Vision-Language-Action (VLA) models connect vision-language pre-training with robot control. RT-2~\cite{RT2_2023_CoRL} and OpenVLA~\cite{OpenVLA_2024_CoRL} transfer VLM semantic priors to action generation, while FAST~\cite{FAST_2025_arXiv} improves efficiency through frequency-space compression. For continuous multi-step control, Octo~\cite{Octo_2024_arXiv}, $\pi_0$~\cite{PI0_2024_arXiv}, and RDT-1B~\cite{RDT1B_2025_ICLR} use diffusion or flow-matching action heads, and $\pi_{0.5}$~\cite{PI05_2025_arXiv} scales this direction with heterogeneous data and multimodal supervision. Other work addresses robot-data scarcity with synthetic data or human videos~\cite{GR00T_2025_arXiv,Human-as-Humanoid_2026_arXiv,univla_2025_arxiv}, uses structure-aware demonstration selection for data-efficient imitation learning~\cite{SIEVE_2026_arXiv}, improves cross-embodiment transfer through prompt-based adaptation~\cite{HRDT_2025_arXiv,X-VLA_2025_arXiv}, strengthens spatial grounding with 3D-aware representations~\cite{Spatialvla_2025_arXiv,3d_vla_2024_ICML,bridgevla_2025_NearIPS,Evo-0_2025_arxiv,spatialForcing_2025_arXiv,pointvla_2026_RA_L}, or adds world-model, future-prediction, and memory mechanisms for longer-horizon reasoning~\cite{VideoVLA_2025_NeurIPS,VLA-JEPA_2026_arXiv,Memoryvla_2025_arXiv,Mem_0_2026_arXiv}.

\subsection{Intent-based VLA Models}
Intent-driven VLAs aim to bridge semantic reasoning and low-level control. DIAL~\cite{DIAL_2026_arXiv} uses a differentiable latent intent bottleneck with visual foresight, and ACoT-VLA~\cite{ACoT-VLA_2026_arXiv} represents reasoning as kinematically grounded action intents. GazeVLA~\cite{GazeVLA_2026_arXiv} models human intention through gaze as an intermediate representation. MINT~\cite{MINT_2026_arXiv} disentangles low-frequency intent from high-frequency execution residuals, MAIN-VLA~\cite{MAIN_VLA_2026_arXiv} compresses instructions into semantic primitives with structured affordances, DeepVision-VLA~\cite{DeepVision_VLA_2026_arXiv} uses action-guided pruning for task-relevant visual grounding, and VFP~\cite{VFP_2025_arXiv} models multimodal expert behavior with a variational latent prior.
However, these methods rely on the current observation frame, often struggling to resolve short-horizon ambiguity under partial observability where visually similar states require different continuations that can only be disambiguated by recent task history.

\begin{figure*}[t]
    \centering
    \includegraphics[width=1.0\textwidth]{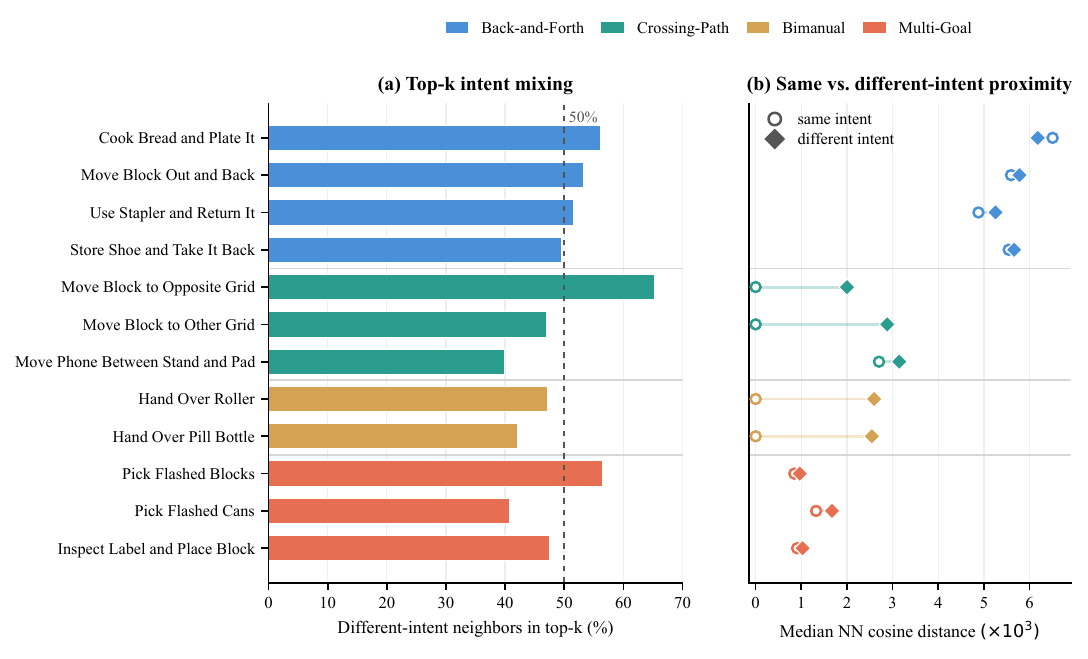}
    \caption{\textbf{Quantitative observation-aliasing diagnostic on \benchmarkname{}.} Embeddings are extracted with Qwen3-VL-8B. Back-and-Forth uses intra-episode retrieval with a 20-frame temporal gap; all other families use cross-episode retrieval. The diagnostic is not a policy success metric. Instead, it measures whether visually nearby states in the ambiguity window can correspond to different next intents. Left: roughly half of the top-$k$ neighbors ($k=5$) come from a different intent. Right: open circles and filled diamonds show median nearest-neighbor cosine distances to same-intent and different-intent states, respectively, with distances scaled by $10^3$ for readability. Although a few tasks appear to have larger distance gaps visually, the actual cosine-distance differences remain on the order of $10^{-3}$.}
    \label{fig:aliasbench_aliasing_diagnostic}
\end{figure*}

\section{\benchmarkname{}: Ambiguity-Aware Benchmark Design}
\label{sec:benchmark_design}

To evaluate whether a policy can resolve aliased observations from recent context, we build \benchmarkname{} on top of RoboTwin2 \cite{RoboTwin2_2025_arXiv}. \benchmarkname{} contains 12 manipulation tasks together with matched simulation training data and held-out evaluation environments. The benchmark targets an underexplored gap in current VLA evaluation: most standard benchmarks measure whether a policy can complete a manipulation task, but they rarely isolate whether the policy can maintain a consistent decision when the current observation is aliased. \benchmarkname{} is therefore designed as a tool for testing whether VLAs can preserve decision consistency across adjacent action chunks. Concretely, we seek task configurations in which two episode states produce nearly identical current observations,
\begin{equation}
o_t^{(1)} \approx o_t^{(2)},
\end{equation}
but require different next actions,
\begin{equation}
a_t^{(1)} \neq a_t^{(2)}.
\end{equation}
The difference should arise from latent context that is not identifiable from the current frame alone but is still recoverable from recent observations. This is the failure mode we want the benchmark to expose.


We group the 12 tasks by the latent factor that resolves the alias: back-and-forth phase, crossing-path source, bimanual handoff direction, and multi-goal cues. These families cover common manipulation patterns in which the current frame is visually plausible for multiple continuations. Figure~\ref{fig:aliasbench_cases} shows representative cases, and Appendix~\ref{sec:appendix_benchmark} provides task-level definitions and examples.

\paragraph{Observation-aliasing diagnostic.}
We quantify whether these aliases are measurable using nearest-neighbor retrieval over visual embeddings inside ambiguity windows; Appendix~\ref{sec:appendix_benchmark} gives the retrieval protocol. The diagnostic is not a policy success metric, but a check that visually nearby states can correspond to different next intents.

Figure~\ref{fig:aliasbench_aliasing_diagnostic} confirms strong current-frame aliasing: the average different-intent neighbor ratio is $49.7\%$, and the largest median same-vs.-different intent gap is below $3\times 10^{-3}$ in cosine distance. We also repeat the same top-5-neighbor diagnostic with PaliGemma 2 3B~\cite{steiner2024paligemma}, obtaining a similar different-intent ratio of $51.3\%$, suggesting that the observed aliasing is not specific to the Qwen3-VL-8B embedding space. Thus, \benchmarkname{} isolates states where the current frame gives weak evidence about the short-horizon intent, while recent history contains the missing context.


\section{Method}
\label{sec:method}


\subsection{Motivation and Problem Formulation}
\label{sec:problem_formulation}

\begin{figure*}[t]
    \centering
    \includegraphics[width=1.0\textwidth]{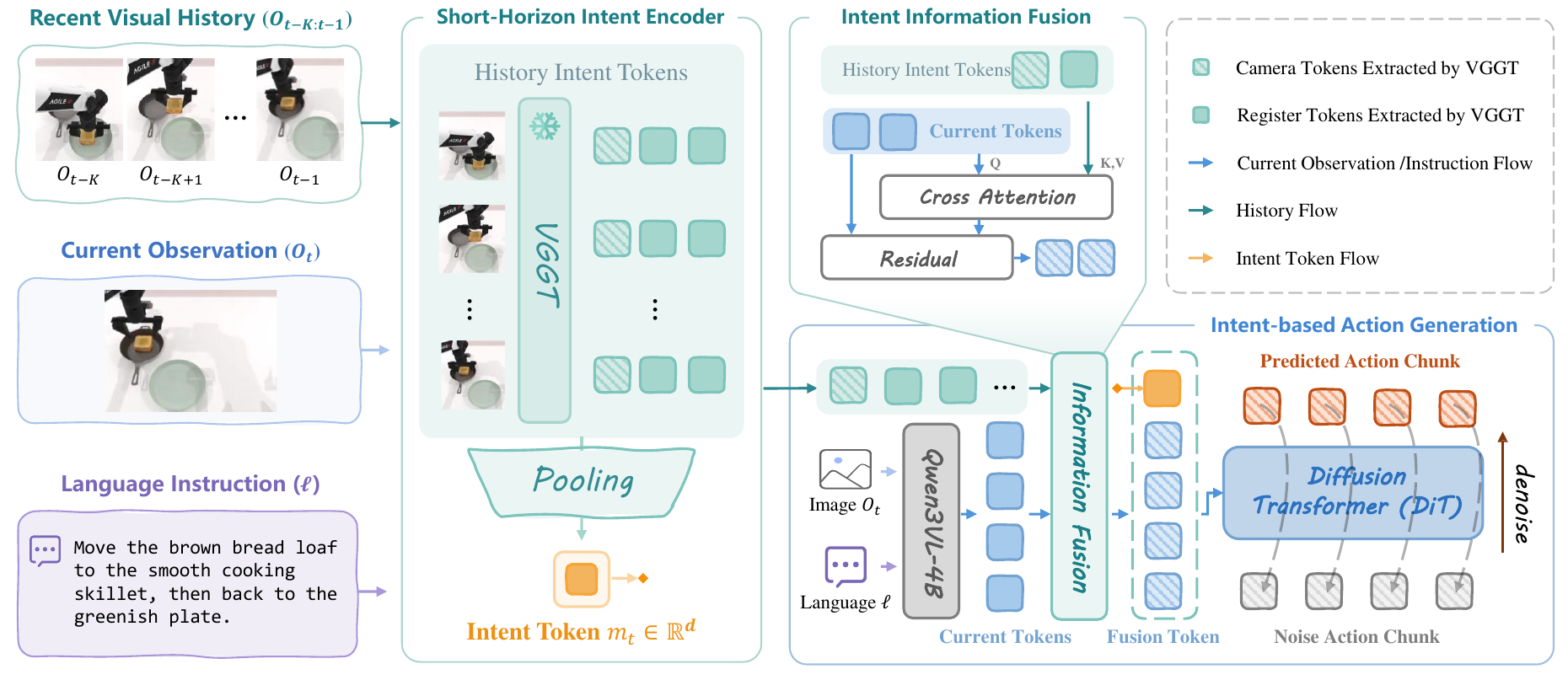}
    \caption{\textbf{Overview of \methodname{}.} A Qwen3-VL backbone encodes the current image and language instruction, while a frozen VGGT-1B history encoder extracts recent visual evidence. \methodname{} fuses the history tokens with the current visual-language context through gated cross-attention, appends a compact short-horizon intent token, and conditions a DiT-based flow-matching action head for chunk generation.}
    \label{fig:method}
\end{figure*}

We model manipulation as a partially observable decision process with latent state $s_t \in \mathcal{S}$, observation $o_t \in \mathcal{O}$, and action $a_t \in \mathcal{A}$. This viewpoint is used only to motivate why recent observations can disambiguate the current frame. At time step $t$, the robot observes $o_t$, receives a language instruction $\ell$, and predicts a future action chunk
\begin{equation}
\tau_t = (a_t, a_{t+1}, \dots, a_{t+H-1}) \in \mathbb{R}^{H \times d_a},
\end{equation}
where $H$ is the chunk horizon and $d_a$ is the action dimension. Instead of using the complete interaction history $H_t=(o_{1:t},a_{1:t-1})$, \methodname{} uses a finite visual history window $h_t^K=o_{t-K:t-1}$ as compact evidence about the recent episode context.
To formalize the ambiguity, let $z_t$ denote a latent short-horizon intent, such as a local continuation mode, task phase, or committed path. A standard frame-conditioned chunk policy models $p_\theta(\tau_t \mid o_t, \ell)$, whose imitation target can be written conceptually as
\begin{equation}
\begin{aligned}
p_\theta(\tau_t \mid o_t, \ell)
= \int &p_\theta(\tau_t \mid o_t, \ell, z_t) \\
&p(z_t \mid o_t, \ell)\,dz_t.
\end{aligned}
\end{equation}
The issue is not multimodality itself, but \emph{uncommitted multimodality under aliased conditioning}: the current frame and instruction may not reveal which continuation has already been selected within the episode. This motivates conditioning chunk generation on recent visual history,
\begin{equation}
\begin{aligned}
p_\theta(\tau_t \mid o_t, \ell, h_t)
= \int &p_\theta(\tau_t \mid o_t, \ell, h_t, z_t) \\
&p(z_t \mid o_t, \ell, h_t)\,dz_t,
\end{aligned}
\end{equation}
where $h_t$ denotes the recent history available at time $t$. Rather than explicitly inferring $z_t$ or supervising intent labels, \methodname{} learns a deterministic short-horizon intent representation $m_t=f_\phi(o_t,\ell,h_t^K)$, which serves as a compact embedding of history-conditioned intent evidence for chunk generation. Throughout the main formulation, $h_t^K$ refers only to recent visual history.
Based on this formulation, we instantiate \methodname{} as shown in Figure~\ref{fig:method}: a frozen visual-history encoder extracts recent intent evidence, a gated fusion module combines this evidence with the current VLA context, and a standard DiT-based flow-matching head generates action chunks. We describe these components below.

\subsection{Short-Horizon Intent from Recent History}
\label{sec:intent_token}

\methodname{} separates the current visual-language context from recent visual history.
In our implementation, the current image and language instruction are processed by a Qwen3-VL 4B backbone $q_\psi$, whose last hidden layer gives the current-condition representation $F_t=q_\psi(o_t,\ell)\in\mathbb{R}^{N\times d}$, where $N$ is the number of current-context tokens and $d$ is the hidden dimension. $F_t$ serves as the conditioning source for the action model.

In parallel, a visual history encoder processes the finite observation history window and produces both history evidence tokens and a summary representation:
\begin{equation}
\begin{aligned}
U_t &= g_\phi(h_t^K) \in \mathbb{R}^{M \times d_h},\\
\bar e_t &= \operatorname{Pool}(U_t) \in \mathbb{R}^{d_h}.
\end{aligned}
\end{equation}
Here $g_\phi$ is the history encoder and operates on image observations. In our method, we instantiate $g_\phi$ with a frozen VGGT-1B encoder \cite{VGGT_2025_CVPR}. When each robot observation contains multiple camera views, the recent-history branch uses only the head-camera frames; the current visual-language backbone can still receive the standard current observation used by the base VLA. 
Specifically, we do not use all VGGT output tokens. Instead, for each input frame, we retain only this one camera token and these four register tokens. The camera token is used by VGGT for camera-parameter prediction, while the register tokens capture global geometric information and inter-frame relations. We use these tokens because they represent recent viewpoint changes and frame-to-frame structure that are particularly useful for inferring the currently active short-horizon intent. The resulting history features are then projected into the action-model hidden space:
\begin{equation}
\tilde U_t = \operatorname{LN}(W_h U_t), \quad
e_t = W_e \bar e_t,
\end{equation}
where $W_h$ and $W_e$ are learned projections and $e_t \in \mathbb{R}^{d}$ is a compact history-evidence token.

Accordingly, the method uses two complementary forms of history information: a sequence of fine-grained history tokens $\tilde U_t$ for token-level fusion, and a single compact token $e_t$ that summarizes recent visual evidence. The compact token is not meant to be a standalone latent intent variable. It provides history evidence, while the condition-dependent intent representation is formed only after this evidence is combined with the current image-language context. All components are learned jointly with the policy objective and require no explicit supervision on intent labels.

\subsection{Intent-based Action Generation}
\label{sec:intent_conditioned_generation}
\label{sec:training_objective}

We fuse the current visual-language context $F_t$~with the history tokens using gated cross-attention. Specifically,
\begin{equation}
\begin{aligned}
F_t' = F_t + \sigma(\alpha)\,\mathrm{MHA}(
&Q=\mathrm{LN}(F_t),\\
&K=\tilde U_t,V=\tilde U_t),
\end{aligned}
\end{equation}
where $\alpha$ is a learned scalar gate and $\mathrm{MHA}$ denotes multi-head attention. The resulting tokens $F_t'$ represent the current observation after it has been enriched with recent history that indicates the active short-horizon continuation.

We also append the projected history-evidence summary as a single context token:
\begin{equation}
e_t^{\mathrm{tok}}=\operatorname{reshape}(e_t) \in \mathbb{R}^{1 \times d}, \quad
C_t = [F_t'; e_t^{\mathrm{tok}}] .
\end{equation}
Conceptually, the condition-dependent information represented by $C_t$ is the learned short-horizon intent representation $m_t=f_\phi(o_t,\ell,h_t^K)$ introduced in Section~\ref{sec:problem_formulation}. In implementation, this representation is realized by the current tokens after gated history fusion together with the appended history-evidence token.
Following the DiT-based conditional flow-matching action heads used in~\cite{starvla_2025,PI05_2025_arXiv,GR00T_2025_arXiv}, we use $C_t$ as the conditioning context for chunk generation. At inference time, $C_t$ is fixed for the current decision step, and the action chunk is obtained by starting from Gaussian noise and integrating the predicted conditional velocity field with the same Euler-style solver used in GR00T.

Training follows the standard conditional flow-matching objective. Given a target action chunk $\tau_t$, Gaussian noise $\epsilon \sim \mathcal{N}(0, I)$, and a sampled flow time $s \sim p(s)$, we define the interpolated chunk $X_s = (1-s)\epsilon + s\tau_t$ and train the conditional velocity field $\hat V_\theta(X_s, s \mid C_t)$ to match the ground-truth displacement $\tau_t - \epsilon$:
\begin{equation}
\mathcal{L}_{\mathrm{flow}}
= \mathbb{E}_{\mathcal{D},\epsilon,s}
\!\left[
\left\|\hat V_\theta(X_s, s \mid C_t)
- (\tau_t - \epsilon)\right\|_2^2
\right].
\end{equation}

\section{Experiment}
\label{sec:experiment}

We begin with \benchmarkname{}, which directly tests the failure mode identified in the introduction. On this benchmark, we compare against Qwen3-VL-GR00T and several history-as-extra-context baselines that feed multiple past frames directly into the Qwen backbone. We then evaluate on SimplerEnv~\cite{SimplerEnv_2024_CoRL}, LIBERO~\cite{LIBERO_2023_NeurIPS}, RoboCasa-GR1 Tabletop Tasks~\cite{RoboCasa_2024_RSS, GR00T_2025_arXiv}, and a real-world snack-cleanup task to test whether the same design transfers beyond the controlled ambiguity benchmark. Across all experiments, we focus on partially observed scenarios where one-frame conditioning is insufficient and analyze both success rate and rollout stability.

\subsection{Results on \benchmarkname{}}
\label{sec:aliasbench_results}

For \benchmarkname{}, we sample 100 demonstrations per task and train all attempted methods under the same 30K-step, 16-H100 budget. Evaluation follows the default RoboTwin2 test protocol~\cite{RoboTwin2_2025_arXiv}; task-specific maximum test steps are listed in Appendix~\ref{sec:appendix_benchmark}. Table~\ref{tab:aliasbench_results} shows that direct raw-history conditioning is costly: the 8-frame and 16-frame variants run out of memory, and the best feasible variant reaches 28.1\% average success. MemoryVLA~\cite{Memoryvla_2025_arXiv} improves over the frame-only Qwen3-VL-GR00T baseline (14.9\% vs. 9.0\%), but remains below the sampled-history baseline. One possible reason is related to memory consolidation. MemoryVLA stores perceptual and cognitive entries in a memory bank and, when the number of entries exceeds the memory length, merges the most similar adjacent entries in each stream by averaging their vectors. This is effective for reducing redundant memories, but \benchmarkname{} is designed so that visually similar states can correspond to different short-horizon intents; Figure~\ref{fig:aliasbench_aliasing_diagnostic} shows that same-intent and different-intent neighbors can be extremely close in feature space. In this setting, similarity-based consolidation may average entries that are visually close but intent-distinct, making the resulting memory less discriminative for selecting the episode-specific continuation. \methodname{} reaches 45.8\%, improving over Qwen3-VL-GR00T by 36.8 points and over the strongest feasible history-as-context baseline by 17.7 points. Gains are largest on crossing-path and back-and-forth tasks, where recent visual history reveals the object source or local phase. Bimanual and multi-goal tasks remain challenging; Appendix~\ref{sec:appendix_icc} further analyzes inter-chunk consistency.

\begin{table}[t]
    \centering
    \footnotesize
    \setlength{\tabcolsep}{3pt}
    \renewcommand{\arraystretch}{1.15}
    \caption{\textbf{Results on \benchmarkname{}.} We compare \methodname{} against Qwen3-VL-GR00T~\cite{starvla_2025}, direct history-as-context baselines, and the memory-centric MemoryVLA baseline~\cite{Memoryvla_2025_arXiv}. `OOM' means the corresponding training configuration runs out of GPU memory when past frames are fed directly into the Qwen backbone as extra context.}
    \label{tab:aliasbench_results}
     \resizebox{\linewidth}{!}{
    \begin{tabular}{lccccc}
        \toprule
        \textbf{Method} & \makecell[c]{\textbf{Back}\\ \textbf{Forth}} & \makecell[c]{\textbf{Crossing}\\ \textbf{Path}} & \makecell[c]{\textbf{Bi-}\\ \textbf{manual}} & \makecell[c]{\textbf{Multi}\\ \textbf{Goal}} & \textbf{Avg} \\
        \midrule
        Qwen3-VL-GR00T \cite{starvla_2025} & 6.0 & 15.7 & 5.5 & 8.7 & 9.0 \\
        + last 16 history frames & OOM & OOM & OOM & OOM & OOM \\
        + last 8 history frames & OOM & OOM & OOM & OOM & OOM \\
        + last 4 history frames & 7.3 & 19.3 & 2.5 & 11.0 & 10.4 \\
        + 4 frames uniformly sampled from last 16 & 31.8 & 47.3 & 6.0 & 18.7 & 28.1 \\
        MemoryVLA~\cite{Memoryvla_2025_arXiv} & 13.3 & 22.7 & 4.0 & 16.7 & 14.9 \\
        \rowcolor{gray!20}
        \methodname{} & \textbf{49.3} & \textbf{74.7} & \textbf{17.0} & \textbf{31.3} & \textbf{45.8} \\
        \bottomrule
    \end{tabular}}
\end{table}

\subsection{Results on Standard Benchmarks}
\label{sec:main_results}

To evaluate whether the advantage of \methodname{} transfers beyond the controlled aliases in \benchmarkname{}, we further conduct experiments on the SimplerEnv, LIBERO, and RoboCasa. Appendix~\ref{sec:appendix_training} provides the dataset and training details.

\definecolor{navyblue}{HTML}{0071BC}

\begin{table}[!t]
  \centering
  \caption{
    \textbf{Results of evaluating the VLA models with the WidowX robot in the SimplerEnv simulation environment~\cite{SimplerEnv_2024_CoRL}}. We highlight the best results in \textbf{bold} and the second-best results with \underline{underline}.
    }
  \resizebox{\linewidth}{!}{
  \rowcolors{2}{gray!15}{white}
  \begin{tabular}{l c c c c c}
    \toprule
    \textbf{Method}
     & \makecell[c]{\textbf{Stack} \\ \textbf{Block}} 
     & \makecell[c]{\textbf{Put} \\ \textbf{Carrot}} 
     & \makecell[c]{\textbf{Put} \\ \textbf{Spoon}} 
     & \makecell[c]{\textbf{Put} \\ \textbf{Eggplant}} 
     & \textbf{Avg} \\
    \midrule
    RT-1-X~\cite{OXE_2024_ICRA}         &  0.0  & 4.2   & 0.0   & 0.0   & 1.1 \\
    Octo-Base~\cite{Octo_2024_arXiv}      & 0.0   & 12.5  & 15.8  & 41.7  & 17.5 \\
    Octo-Small~\cite{Octo_2024_arXiv}     & 0.0   & 8.2   & 41.7  & 56.7  & 26.7 \\
    OpenVLA-OFT~\cite{OpenVLA-OFT_2025_arXiv}    & 30.0  & 30.0  & 34.2  & 72.5  & 41.8 \\
    RoboVLM~\cite{RoboVLM_2024_arXiv}        & 0.0   & 37.5  & 50.0  & 83.3  & 42.7 \\ 
    Magma~\cite{yang2025magma}          & 20.8  & 29.2  & 37.5  & 91.7  & 44.8 \\  
    CogACT~\cite{CogACT_2024_arXiv}         & 15.0 &  50.8  & 71.7 & 67.5 & 51.3 \\
    SpatialVLA~\cite{Spatialvla_2025_arXiv}     & 25.0  & 20.8  & 20.8  & 70.8  & 34.4 \\
    TraceVLA~\cite{TraceVLA_2025_arXiv}       & 16.6  & 16.6  & 12.5  & 65.0  & 27.7 \\
    VideoVLA~\cite{VideoVLA_2025_NeurIPS}       & 45.8 & 20.8   & 75.0 & 70.8 & 53.1 \\
    $\pi_0$~\cite{PI0_2024_arXiv}                   & 29.2 & 62.5 & 29.2 & 91.6 & 53.1 \\
    $\pi_{0.5}$~\cite{PI05_2025_arXiv}               & 44.7 & 64.7 & 49.3 & 69.7 & 57.1 \\
    GR00T-N1.6~\cite{GR00T_N1.6}   & 5.5 & 65.5 & 64.5 & 93.0 & 57.1 \\
    LangForce~\cite{LangForce_2026_arXiv} & 33.3 &  63.8 & 89.6 & 79.2 & 66.5 \\
    PhysBrain~\cite{PhysBrain_2025_arXiv} & 34.7 & 58.3 & 90.3 & 80.6 & 65.9 \\
    3D-Mix~\cite{3DMix_2026_arXiv} & 30.2 & 61.5 & 86.5  & 94.8 & 68.2 \\
    TwinBrainVLA~\cite{TwinBrainVLA_2026_arXiv}  &   33.3 &  58.3   &  87.5  &  79.1 &  64.5 \\
    MemoryVLA~\cite{Memoryvla_2025_arXiv} & 37.5 & 75.0 & 75.0 & 100.0 & 71.9 \\
    \midrule
     Qwen3-VL-GR00T~\cite{starvla_2025}  &  18.8 &  59.4 &  83.0 & 100.0 &  65.3 \\
    \rowcolor{gray!30} \textbf{\methodname}  &  54.2 & 66.7 & 70.8 & 100.0 & \textbf{72.9} \\
    \bottomrule
  \end{tabular}}
  \vspace{-0.5em}
  \label{tab:simplerenv}
\end{table}

\begin{table}[!t]
    \centering
    \small
    \caption{\textbf{Comparison on the LIBERO benchmark.} We train one policy for all 4 suites.}
    \label{tab:libero_results}
    \renewcommand{\arraystretch}{1.15} 
    \setlength{\tabcolsep}{3pt} 
    \resizebox{\linewidth}{!}{
    \rowcolors{2}{gray!15}{white}
    \begin{tabular}{lccccc}
        \toprule
        \textbf{Method} & \textbf{Spatial} & \textbf{Object} & \textbf{Goal} & \textbf{Long} & \textbf{Avg} \\
        \midrule
        OpenVLA~\cite{OpenVLA_2024_CoRL} & 87.4 & 88.4 & 79.2 & 53.7 & 76.5 \\
        OpenVLA-OFT~\cite{OpenVLA-OFT_2025_arXiv} & 97.6 & 98.4 & 97.9 & 94.5 & 97.1 \\
        $\pi_0$~\cite{PI0_2024_arXiv} & 96.8 & 98.8 & 95.8 & 85.2 & 94.1 \\
        $\pi_{0.5}$~\cite{PI05_2025_arXiv} & 98.8 & 98.2 & 98.0 & 92.4 & 96.9 \\
        VLA-JEPA~\cite{VLA-JEPA_2026_arXiv}  & 94.8 & 99.6 & 95.8 & 94.0 & 96.1 \\
        TwinBrainVLA~\cite{TwinBrainVLA_2026_arXiv} & 99.2 & 99.0 & 96.8 & 95.4 & 97.6 \\
        \midrule
        Qwen3-VL-GR00T~\cite{starvla_2025} & 97.8 & 98.8 & 97.4 & 92.0 & 96.5 \\
        \rowcolor{gray!30}\textbf{\methodname{}} & 99.3 & 99.7 & 98.1 & 97.4 & \textbf{98.6} \\
        \bottomrule
    \end{tabular}
    }
\end{table}


\paragraph{Results on SimplerEnv.}
For SimplerEnv~\cite{SimplerEnv_2024_CoRL}, we evaluate on four WidowX manipulation tasks: \emph{Put Spoon on Towel}, \emph{Put Carrot on Plate}, \emph{Stack Green Block on Yellow Block}, and \emph{Put Eggplant in Yellow Basket}. The results are reported in Table~\ref{tab:simplerenv}. \methodname{} achieves the best overall average success rate of 72.9\%, outperforming the Qwen3-VL-GR00T baseline by 7.6 points and exceeding the strongest previously reported average, 68.2\% from 3D-Mix, by 4.7 points. The gains are especially large on \emph{Put Carrot on Plate}, \emph{Stack Green Block on Yellow Block}, and \emph{Put Eggplant in Yellow Basket}. Although performance on \emph{Put Spoon on Towel} is lower than the baseline, the overall result shows that recent visual history substantially improves robustness on partially observed chunked manipulation.

\paragraph{Results on LIBERO.}
For LIBERO~\cite{LIBERO_2023_NeurIPS}, we report Avg@500 success rates on Spatial, Object, Goal, and Long. As shown in Table~\ref{tab:libero_results}, LIBERO is already close to saturation for strong recent VLAs, especially on the Spatial, Object, and Goal suites where several methods exceed 98\%. Therefore, these suites leave limited room to diagnose the effect of short-horizon history. We focus on LIBERO-Long, which is the most relevant suite for our setting because it contains longer, multi-stage manipulation routines where adjacent local continuations must remain consistent. On LIBERO-Long, \methodname{} reaches 97.4\%, compared with 92.0\% for the Qwen3-VL-GR00T baseline and 92.4\% for $\pi_{0.5}$. Although \methodname{} does not introduce an explicit long-horizon planner, recent visual history helps it infer the active local continuation inside a longer routine, which reduces inconsistent chunk generation across sub-steps.

\begin{table}[!t]
    \centering
    \small
    \setlength{\tabcolsep}{3.5pt}
    \renewcommand{\arraystretch}{1.15}
    \caption{
    RoboCasa-GR1 simulation results on four representative pick-and-place tasks.
    The omitted columns indicate additional RoboCasa-GR1 tasks; Avg. is computed over all 24 tasks rather than only the shown tasks.
    Full per-task results are provided in Appendix Table~\ref{tab:robocasa_full_results}.
    }
    \label{tab:robocasa_main_tab}
    \resizebox{\linewidth}{!}{
    \rowcolors{2}{gray!15}{white}
    \begin{tabular}{lcccccc}
        \toprule
        \textbf{Method} & \makecell[c]{\textbf{PnP} \\ \textbf{Bottle}} & \makecell[c]{\textbf{PnP} \\ \textbf{Can}} & \makecell[c]{\textbf{PnP} \\ \textbf{Cup}} & \makecell[c]{\textbf{PnP} \\ \textbf{Milk}} & $\cdots$ & \textbf{Avg.} \\
        \midrule
        GR00T N1.5~\cite{GR00T_2025_arXiv}      & 54.0 & 50.0 & 38.0 & 60.0 & $\cdots$ & 48.2 \\
        GR00T N1.6~\cite{GR00T_N1.6}            & 51.5 & 13.0 &  8.5 & 14.0 & $\cdots$ & 47.6 \\
        VP-VLA~\cite{VP-VLA_2026_arXiv}         & 54.0 & 72.0 & 44.0 & 74.0 & $\cdots$ & 53.8 \\
        TwinBrainVLA~\cite{TwinBrainVLA_2026_arXiv} & 74.0 & 72.0 & 52.0 & 60.0 & $\cdots$ & \underline{54.6} \\
        PhysBrain~\cite{PhysBrain_2025_arXiv}   & 74.0 & 68.0 & 42.0 & 54.0 & $\cdots$ & 50.0 \\
        LangForce~\cite{LangForce_2026_arXiv}   & 72.0 & 78.0 & 46.0 & 56.0 & $\cdots$ & 52.6 \\
        \midrule
        Full-Frame Training~\cite{starvla_2025} & 46.0 & 80.0 & 54.0 & 48.0 & $\cdots$ & 47.8 \\
        \textbf{\methodname{} (ours)}           & 76.0 & 88.0 & 46.0 & 48.0 & $\cdots$ & \textbf{57.0} \\
        \bottomrule
    \end{tabular}}
\end{table}

\paragraph{Results on RoboCasa.}
We evaluate \methodname{} on the RoboCasa GR1 Tabletop Manipulation Benchmark \cite{RoboCasa_2024_RSS,GR00T_2025_arXiv}, which contains 24 diverse manipulation tasks with articulated objects and varied object geometries. The benchmark includes tasks such as \emph{PnPBottleToCabinetClose} and \emph{PnPCanToDrawerClose}, as well as scenarios involving appliances such as microwaves and toasters. As shown in Table~\ref{tab:robocasa_main_tab}, \methodname{} achieves the best 24-task average success rate of 57.0\%, outperforming TwinBrainVLA (54.6\%), VP-VLA (53.8\%), LangForce (52.6\%), and Qwen3-VL-GR00T (47.8\%). Full per-task results are provided in Appendix~\ref{sec:appendix_robocasa}. These results indicate that the short-horizon history signal remains useful even in a broader benchmark with articulated objects and varied interaction patterns.

\subsection{Real-World Experiment}
\label{sec:realworld_snack}

We evaluate \methodname{} on a real-world snack-cleanup task using two Franka Research 3 arms with Robotiq 85 grippers and Intel RealSense D435 head/hand cameras. Each episode starts with four randomly placed snacks, and the robot must place them into a bag within 500 action chunks. We train from 300 full four-snack demonstrations. The task creates multi-object ambiguity: several snacks and the bag remain visible, so a frame-only policy can start toward one snack but switch to another after replanning. Real camera noise, occlusion, and small execution deviations further test whether the policy preserves a stable target commitment. Table~\ref{tab:realworld_snack} compares \methodname{} with Qwen3-VL-GR00T, $\pi_{0.5}$, and MemoryVLA over 50 trials per model. \methodname{} achieves the highest cumulative rate at every threshold and the highest expected count, reaching 1.86 compared with 1.20, 1.44, and 1.54 for Qwen3-VL-GR00T, $\pi_{0.5}$, and MemoryVLA, respectively.

\begin{table}[t]
    \centering
    \small
    \setlength{\tabcolsep}{3pt}
    \renewcommand{\arraystretch}{1.3}
    \caption{
    \textbf{Real-world Experiment on a dual-arm Franka Research 3 setup.}
    Each rollout starts with four randomly placed snacks on the table and runs for a 500-action-chunk budget.
    Columns $\ge$1--$\ge$4 are cumulative threshold counts: the number of 50 trials in which at least that many snacks are placed into the bag, so they are not disjoint. $\mathbb{E}[\#]$ is the expected number of placed snacks per rollout.
    }
    \label{tab:realworld_snack}
    \resizebox{\linewidth}{!}{
    \begin{tabular}{lccccc}
        \toprule
        \textbf{Method} & \textbf{$\ge$1} & \textbf{$\ge$2} & \textbf{$\ge$3} & \textbf{$\ge$4} & \textbf{$\mathbb{E}[\#]$} \\
        \midrule
        Qwen3-VL-GR00T~\cite{starvla_2025} & 35/50 (70\%) & 18/50 (36\%) & 7/50 (14\%) & 0/50 (0\%) & 1.20 \\
        $\pi_{0.5}$~\cite{PI05_2025_arXiv} & 41/50 (82\%) & 22/50 (44\%) & 9/50 (18\%) & 0/50 (0\%) & 1.44 \\
        MemoryVLA~\cite{Memoryvla_2025_arXiv} & 39/50 (78\%) & 26/50 (52\%) & 11/50 (22\%) & 1/50 (2\%) & 1.54 \\
        \rowcolor{gray!30}\textbf{\methodname{} (ours)} & 43/50 (86\%) & 31/50 (62\%) & 16/50 (32\%) & 3/50 (6\%) & 1.86 \\
        \bottomrule
    \end{tabular}}
\end{table}

\subsection{Ablation Studies}
\label{sec:ablations}

\begin{table}[!t]
  \centering
  \caption{
    \textbf{Ablation study on SimplerEnv.} We ablate the history encoder, temporal history, history fusion, and the compact intent-evidence token. Results are success rates on the four WidowX tasks.
  }
  \label{tab:simplerenv_ablation}
  \small
  \setlength{\tabcolsep}{3.0pt}
  \renewcommand{\arraystretch}{1.15}
  \resizebox{\linewidth}{!}{
  \begin{tabular}{lccccc}
    \toprule
    \textbf{Variant}
     & \makecell[c]{\textbf{Stack} \\ \textbf{Block}} 
     & \makecell[c]{\textbf{Put} \\ \textbf{Carrot}} 
     & \makecell[c]{\textbf{Put} \\ \textbf{Spoon}} 
     & \makecell[c]{\textbf{Put} \\ \textbf{Eggplant}} 
     & \textbf{Avg} \\
    \midrule
    Frame-only Qwen3-VL-GR00T & 18.8 & 59.4 & \textbf{83.0} & \textbf{100.0} & 65.3 \\
    VGGT, current frame only & 30.2 & 61.5 & \underline{72.5} & 94.8 & 64.8 \\
    History fusion only, no intent token & \underline{49.0} & \underline{65.6} & 67.7 & 95.8 & \underline{69.5} \\
    \rowcolor{gray!30}
    \methodname{} & \textbf{54.2} & \textbf{66.7} & 70.8 & \textbf{100.0} & \textbf{72.9} \\
    \bottomrule
  \end{tabular}}
\end{table}

\paragraph{Component ablations on SimplerEnv.}
Table~\ref{tab:simplerenv_ablation} studies which components drive the gains on SimplerEnv. Removing VGGT and temporal history gives the frame-only Qwen3-VL-GR00T baseline. Adding VGGT to only the current frame does not improve the average score, showing that VGGT is not useful merely as another single-frame encoder; its benefit comes from geometric and inter-frame evidence in a recent window. History fusion raises the average from 65.3\% to 69.5\%, and the compact intent-evidence token further raises it to 72.9\%, indicating that both temporal fusion and an explicit history summary help chunk generation.

The per-task results clarify where history helps. \emph{Put Spoon on Towel} drops relative to the frame-only baseline because it has little short-horizon intent ambiguity: once the spoon and towel are visible, the current frame already carries most needed information, and history may dilute attention to the small, thin spoon. In contrast, \emph{Stack Green Block on Yellow Block} improves from 18.8\% to 54.2\%, as stacking depends on relative 3D geometry, alignment, and the grasp-to-placement transition, where recent visual history and VGGT geometry tokens provide useful cues.

\section{Conclusion}
\label{sec:conclusion}

We introduced \methodname{}, a history-conditioned VLA for stabilizing chunk generation under partial observability, and \benchmarkname{}, a 12-task benchmark for short-horizon observation aliasing. Across \benchmarkname{}, standard simulation benchmarks, and a real-world cleanup task, compact recent-history conditioning improves success and inter-chunk consistency.

\section*{Limitations}

\methodname{} focuses on recovering short-horizon intent from recent visual history. The history pathway is used only as an auxiliary signal for intent disambiguation during chunk generation; we do not claim general episodic memory or long-horizon memory as a contribution. Accordingly, our intended application scope is locally aliased manipulation, where recent frames reveal the active continuation but the current frame alone is ambiguous. Tasks that require remembering sparse events outside the recent window or recovering from large closed-loop deviations are outside this scope and may need longer-term memory or planning modules.   


\section*{Ethical Considerations}

This work studies short-horizon intent disambiguation for VLA policies in controlled simulation benchmarks and a small lab robot setup. It does not use personal, sensitive, or offensive language data. The real-world demonstrations consist of robot observations and actions collected by expert teleoperation, without identifying human information. The main ethical risk is misinterpretation: high benchmark scores should not be treated as evidence that a policy is safe or reliable for open-world physical deployment, and low scores are diagnostic signals under our aliasing setting, not a complete model characterization. We intend the benchmark, code, and model weights for research evaluation, with standard physical-robot safety practices, human supervision, and further robustness validation before deployment.

\section*{Acknowledgments}

This work is supported in part by the Beijing Zhongguancun Academy (Grant No. C20250510); in part by the National Natural Science Foundation of China under Grant U23A20300; and in part by the High Innovation Plan (Grant No. 202504841022).

\bibliography{custom}

\newpage

\appendix



\section{Training and Data Details}
\label{sec:appendix_training}

Unless otherwise specified, the standard-benchmark experiments are built on the StarVLA training pipeline~\cite{starvla_2025} and run on 16 NVIDIA H100 GPUs. We follow the default StarVLA training protocol for fair comparison, and use AdamW~\cite{AdamW_2017_ICLR} with an initial learning rate of $1\times 10^{-5}$ and a cosine annealing schedule. System-level optimizations include DeepSpeed ZeRO-2~\cite{rasley2020deepspeed}, gradient clipping with maximum norm 1.0, and no gradient accumulation.

\paragraph{SimplerEnv.}
We use the BridgeDataV2~\cite{Bridgedatav2_2023_CoRL} subset of Open X-Embodiment~\cite{OXE_2024_ICRA} and fine-tune the model for 30K steps on 16 GPUs with batch size 16 per device. We evaluate the policy with the official SimplerEnv evaluation scripts~\cite{SimplerEnv_2024_CoRL}.

\paragraph{LIBERO.}
We train a single policy jointly across the four LIBERO suites and evaluate Spatial, Object, Goal, and Long.

\paragraph{RoboCasa-GR1.}
For RoboCasa-GR1, we use the Humanoid Robot Tabletop Manipulation subset of PhysicalAI Robotics-GR00T-X-Embodiment-Sim~\cite{GR00T_2025_arXiv} and fine-tune the model for 30K steps on 16 GPUs with batch size 16 per device. 

\section{Additional Ablation}
\label{sec:appendix_additional_ablation}

\subsection{History-Window Length}
\label{sec:appendix_history_window}

To further analyze the \methodname{} result in Table~\ref{tab:aliasbench_results}, we ablate the history-window length $K$. At 30 FPS, $K=8$, $16$, and $24$ cover approximately $0.27$, $0.53$, and $0.80$ seconds, respectively. Our practical principle is to choose the shortest window that covers the key disambiguating event---such as a phase transition, pickup origin, or transient cue---rather than simply maximize the number of frames. The importance of temporal coverage is also visible in Table~\ref{tab:aliasbench_results}: uniformly sampling four frames from the last 16 raises the average from 10.4\% to 28.1\% relative to the last four consecutive frames. Table~\ref{tab:history_window} shows that $K=16$ gives the best overall result. Multi-Goal improves with a longer window, while $K=24$ slightly reduces Back-and-Forth and Bimanual performance, reflecting a trade-off between retaining earlier cues and introducing stale evidence. Compared with the 28.1\% sparse raw-history baseline covering the same 16-frame interval, the 45.8\% result at $K=16$ further suggests that temporal coverage alone does not explain the full gain.

\begin{table}[!t]
    \centering
    \footnotesize
    \setlength{\tabcolsep}{5pt}
    \renewcommand{\arraystretch}{1.15}
    \caption{\textbf{History-window length on \benchmarkname{}.}}
    \label{tab:history_window}
    \resizebox{\linewidth}{!}{
    \begin{tabular}{lccccc}
        \toprule
        \textbf{Window} & \textbf{Back Forth} & \textbf{Crossing Path} & \textbf{Bimanual} & \textbf{Multi Goal} & \textbf{Avg} \\
        \midrule
        $K=8$  & 45.6 & 73.1 & 12.9 & 27.8 & 39.9 \\
        $K=16$ & \textbf{49.3} & 74.7 & \textbf{17.0} & 31.3 & \textbf{45.8} \\
        $K=24$ & 47.2 & \textbf{75.1} & 16.4 & \textbf{35.9} & 43.7 \\
        \bottomrule
    \end{tabular}}
\end{table}

\subsection{Alternative History Encoders}
\label{sec:appendix_history_encoder}

To test whether the history-conditioning framework depends on VGGT, we replace it with the video-based V-JEPA 2 encoder~\cite{VJEPA2_2025_arXiv} while keeping the remaining framework unchanged. As shown in Table~\ref{tab:history_encoder}, V-JEPA 2 reaches a 38.6\% average, showing that a video encoder can also compress recent history into useful temporal evidence. VGGT performs better across all four families and improves the average by 7.2 points. This advantage is consistent with VGGT's multi-view geometric pre-training and alternating frame-wise and global attention, which capture both within-frame content and cross-frame structure~\cite{VGGT_2025_CVPR}.

\begin{table}[!t]
    \centering
    \footnotesize
    \setlength{\tabcolsep}{5pt}
    \renewcommand{\arraystretch}{1.15}
    \caption{\textbf{Alternative history encoders on \benchmarkname{}.}}
    \label{tab:history_encoder}
    \resizebox{\linewidth}{!}{
    \begin{tabular}{lccccc}
        \toprule
        \textbf{Encoder} & \textbf{Back Forth} & \textbf{Crossing Path} & \textbf{Bimanual} & \textbf{Multi Goal} & \textbf{Avg} \\
        \midrule
        V-JEPA 2 & 41.6 & 71.2 & 13.5 & 27.9 & 38.6 \\
        VGGT & \textbf{49.3} & \textbf{74.7} & \textbf{17.0} & \textbf{31.3} & \textbf{45.8} \\
        \bottomrule
    \end{tabular}}
\end{table}

\subsection{Inference Efficiency}
\label{sec:appendix_efficiency}

We measure BF16 deployment throughput and forward-pass peak GPU memory on a single NVIDIA H100 80GB GPU, using the same batch size and precision for all methods. As shown in Table~\ref{tab:inference_efficiency}, adding the \methodname{} history pathway changes throughput from 11.77 to 7.53 Hz and forward peak memory from 9.83 to 14.58 GiB relative to frame-only Qwen3-VL-GR00T, quantifying the cost of 16-frame history conditioning. Compared with directly processing 16 raw history images, however, \methodname{} is faster (7.53 vs. 6.45 Hz) and uses 5.55 GiB less forward memory (14.58 vs. 20.13 GiB). Thus, compact history encoding is substantially more efficient than direct conditioning on the same history length.

\begin{table}[!t]
    \centering
    \footnotesize
    \setlength{\tabcolsep}{3pt}
    \renewcommand{\arraystretch}{1.15}
    \caption{\textbf{BF16 deployment efficiency on one NVIDIA H100 80GB GPU.}}
    \label{tab:inference_efficiency}
    \resizebox{\linewidth}{!}{
    \begin{tabular}{lcc}
        \toprule
        \textbf{Method} & \textbf{Throughput (Hz) $\uparrow$} & \textbf{Forward Peak (GiB) $\downarrow$} \\
        \midrule
        Qwen3-VL-GR00T & 11.77 & 9.83 \\
        + last 4 history frames & 9.68 & 11.80 \\
        + last 8 history frames & 8.70 & 15.71 \\
        + last 16 history frames & 6.45 & 20.13 \\
        \rowcolor{gray!20}
        \methodname{} & 7.53 & 14.58 \\
        \bottomrule
    \end{tabular}
    }
\end{table}

\section{Additional Analysis on Intent Consistency and Mode Switching}
\label{sec:appendix_icc}

\subsection{Inter-Chunk Consistency in Ambiguous Windows}

\begin{figure*}[t]
    \centering
    \includegraphics[width=1.0\textwidth]{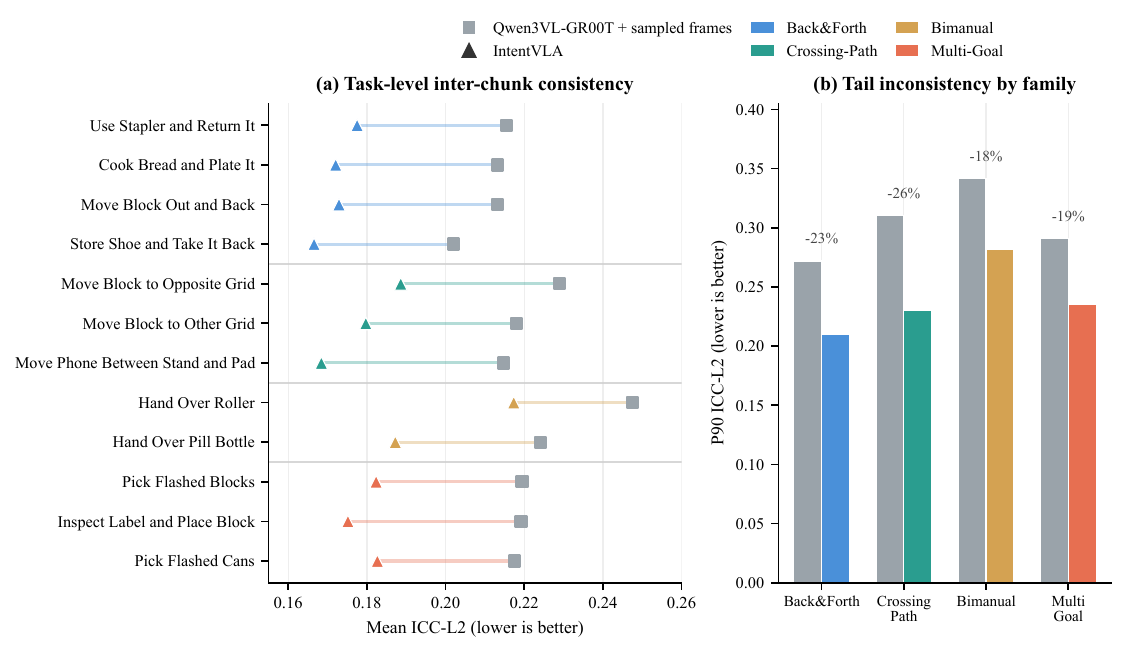}
    \caption{\textbf{Inter-chunk consistency in \benchmarkname{} ambiguity windows.} We compare \methodname{} against the strongest feasible history-as-context baseline in Table~\ref{tab:aliasbench_results}, Qwen3-VL-GR00T with four frames uniformly sampled from the last 16 frames. ICC-L2 is the squared L2 overlap error defined in Eq.~\eqref{eq:icc_l2}; lower values indicate more consistent local intent. Left: task-level mean ICC-L2. Right: family-level 90th-percentile ICC-L2, where the 90th percentile measures tail inconsistency among the harder ambiguity windows. In the right panel, gray bars show the results of Qwen3-VL-GR00T with sampled frames.}
    \label{fig:aliasbench_icc}
\end{figure*}

We further evaluate whether the policy preserves the same local intent across adjacent replanning steps. Consider a chunk $\hat{\tau}^{(t)}=(\hat a_t^{(t)},\ldots,\hat a_{t+H-1}^{(t)})$ generated at decision time $t$, and another chunk $\hat{\tau}^{(t+r)}$ generated after replanning at time $t+r$. These two chunks overlap on future time steps $t+r,\ldots,t+H-1$. We define the inter-chunk consistency error with an L2 action distance:
\begin{equation}
\mathrm{ICC}_t
=
\frac{1}{H-r}
\sum_{j=r}^{H-1}
\left\|
\hat a_{t+j}^{(t)}
-
\hat a_{t+j}^{(t+r)}
\right\|_2^2 .
\label{eq:icc_l2}
\end{equation}
Here $\hat a_{t+j}^{(t)}$ denotes the action for absolute time $t+j$ predicted by the chunk sampled at time $t$, while $\hat a_{t+j}^{(t+r)}$ denotes the prediction for the same absolute time made by the next replanning step. We refer to this metric as ICC-L2. It is computed only inside annotated ambiguity windows in \benchmarkname{}, where the current observation alone does not identify the correct continuation. Lower ICC-L2 is better: it means that adjacent chunks agree more strongly on their overlapping future segment, which is the expected action-level signature of preserved short-horizon intent.

Figure~\ref{fig:aliasbench_icc} shows that \methodname{} reduces inter-chunk inconsistency across all 12 tasks. Averaged over tasks, mean ICC-L2 decreases from $0.219$ to $0.181$, a $17.6\%$ relative reduction. The family-level view also shows lower tail inconsistency across all ambiguity families. These results indicate that recent visual history makes adjacent replanned action chunks more consistent in the ambiguous regions where frame-conditioned chunk policies are likely to change intent.

\subsection{Mode Switching Under Receding-Horizon Sampling}
\label{sec:theory_mode_switch}

The failure mode studied in this paper is not that a policy can represent multiple valid behaviors. The problem is that, under an aliased current observation, the policy may remain \emph{uncommitted} about which local continuation is active. Receding-horizon execution then turns this uncommitted multimodality into a temporal consistency problem: two adjacent action chunks may each be plausible in isolation, but they can correspond to different short-horizon intents. Recent visual history reduces this risk by anchoring both chunks to the same episode-specific continuation.

To make this point explicit, suppose the policy replans every $r$ environment steps. Let $p_t(z)$ and $p_{t+r}(z)$ denote conceptual intent distributions at two adjacent decision steps, where $z\in\mathcal{Z}$ indexes local continuations such as task phase, source-conditioned destination, or handoff direction. These distributions are only used for analysis; \methodname{} does not explicitly infer a discrete intent label. If the two chunks are sampled independently from these intent distributions, the probability that they correspond to different intents is
\begin{equation}
P_{\mathrm{switch}}(t,r)
=
1-\sum_{z\in\mathcal{Z}} p_t(z)\,p_{t+r}(z).
\end{equation}
In an aliased region, the current frame may leave several continuations plausible. If $p_t(z)=p_{t+r}(z)$ is uniform over $M$ plausible intents, then $P_{\mathrm{switch}}(t,r)=1-1/M$: adjacent chunks are likely to switch intent even though neither chunk is individually invalid. By contrast, if recent history concentrates both distributions around the same committed intent $z^\star$, then the switch probability approaches zero. Thus, the role of history is not to remove multimodality across episodes, but to preserve within-episode commitment during local replanning.

The latent intent $z$ is not observed for a learned policy, so we evaluate the consequence of mode switching in action space. When two adjacent chunks predict actions for overlapping absolute timesteps, a switch in short-horizon intent should appear as disagreement between the two predictions. The ICC-L2 metric in Eq.~\eqref{eq:icc_l2} measures exactly this overlap disagreement inside annotated ambiguity windows. A large ICC-L2 therefore indicates a possible action-level manifestation of mode switching, while a small ICC-L2 indicates that adjacent chunks remain aligned with the same continuation. This makes ICC-L2 an observable proxy for the consistency effect implied by $P_{\mathrm{switch}}(t,r)$.

The full ICC statistics in Figure~\ref{fig:aliasbench_icc} support this interpretation. Beyond the mean reduction reported in the main text, the task-averaged 90th-percentile ICC-L2 decreases from $0.298$ to $0.233$, a $21.7\%$ relative reduction, and the standard deviation across ambiguity windows drops from $0.093$ to $0.046$. This indicates that \methodname{} reduces both average overlap disagreement and unstable high-error windows. At the family level, the 90th-percentile ICC-L2 improves consistently across back-and-forth, crossing-path, bimanual, and multi-goal ambiguity, with relative reductions of $22.8\%$, $25.7\%$, $17.5\%$, and $19.2\%$, respectively. These results match the mode-switching analysis: recent visual history makes the effective short-horizon intent conditioning more committed, and adjacent sampled chunks become less likely to follow different continuations in aliased states.

\section{Full RoboCasa-GR1 Results}
\label{sec:appendix_robocasa}

Table~\ref{tab:robocasa_full_results} reports the full RoboCasa-GR1 results corresponding to the average scores in Table~\ref{tab:robocasa_main_tab}.

\begin{table*}[!t]
    \centering
    \footnotesize
    \renewcommand{\arraystretch}{1.47}
    \setlength{\tabcolsep}{5pt}
    \caption{
      \textbf{Full RoboCasa-GR1 Tabletop results.}
      We report Avg@50 success rates (\%) for 24 tasks. Results for Isaac-GR00T N1.5 and N1.6 are sourced from the official Isaac-GR00T repository~\cite{GR00T_2025_arXiv}; Qwen3-VL-GR00T results are sourced from the official StarVLA experiments~\cite{starvla_2025}.
    }
    \label{tab:robocasa_full_results}
    \begin{adjustbox}{width=\textwidth}
    \begin{tabular}{l c c c c c c c c}
        \toprule
        Task &
        {\scriptsize \makecell{\textbf{GR00T}\\\textbf{N1.5}}} &
        {\scriptsize \makecell{\textbf{GR00T}\\\textbf{N1.6}}} &
        {\scriptsize \makecell{\textbf{Qwen3-VL}\\\textbf{GR00T}}} &
        {\scriptsize \makecell{\textbf{VP}\\\textbf{VLA}}} &
        {\scriptsize \makecell{\textbf{TwinBrain}\\\textbf{VLA}}} &
        {\scriptsize \makecell{\textbf{Phys}\\\textbf{Brain}}} &
        {\scriptsize \makecell{\textbf{Lang}\\\textbf{Force}}} &
        {\scriptsize \makecell{\textbf{\methodname}}} \\
        \midrule
        PnP Bottle To Cabinet Close                         & 54.0 & 51.5 & 46.0 & 54.0 & 74.0 & 74.0 & 72.0 & 76.0 \\
        PnP Can To Drawer Close                             & 50.0 & 13.0 & 80.0 & 72.0 & 72.0 & 68.0 & 78.0 & 88.0 \\
        PnP Cup To Drawer Close                             & 38.0 &  8.5 & 54.0 & 44.0 & 52.0 & 42.0 & 46.0 & 46.0 \\
        PnP Milk To Microwave Close                         & 60.0 & 14.0 & 48.0 & 74.0 & 60.0 & 54.0 & 56.0 & 48.0 \\
        PnP Potato To Microwave Close                       & 32.0 & 41.5 & 28.0 & 34.0 & 36.0 & 24.0 & 36.0 & 44.0 \\
        PnP Wine To Cabinet Close                           & 38.0 & 16.5 & 46.0 & 48.0 & 46.0 & 54.0 & 46.0 & 56.0 \\
        \midrule
        \rowcolor{gray!20}\textbf{PnP * to * Close (Avg)}   & 45.3 & 24.2 & 50.3 & 54.3 & \underline{56.7} & 52.7 & 55.7 & \textbf{59.7} \\
        \midrule
        PnP Novel From Cuttingboard To Basket               & 38.0 & 58.0 & 48.0 & 66.0 & 62.0 & 62.0 & 66.0 & 66.0 \\
        PnP Novel From Cuttingboard To Cardboardbox         & 46.0 & 46.5 & 40.0 & 54.0 & 46.0 & 44.0 & 40.0 & 52.0 \\
        PnP Novel From Cuttingboard To Pan                  & 58.0 & 68.5 & 68.0 & 74.0 & 70.0 & 56.0 & 68.0 & 56.0 \\
        PnP Novel From Cuttingboard To Pot                  & 62.0 & 65.0 & 52.0 & 54.0 & 66.0 & 58.0 & 48.0 & 54.0 \\
        PnP Novel From Cuttingboard To Tieredbasket         & 28.0 & 46.5 & 56.0 & 56.0 & 52.0 & 40.0 & 44.0 & 46.0 \\
        \midrule
        \rowcolor{gray!20}\textbf{PnP Novel From Cuttingboard To * (Avg)}
                                                              & 46.4 & 56.9 & 52.8 & \textbf{60.8} & \underline{59.2} & 52.0 & 53.2 & 54.8 \\
        \midrule
        PnP Novel From Placemat To Basket                    & 30.0 & 58.5 & 42.0 & 48.0 & 30.0 & 42.0 & 54.0 & 56.0 \\
        PnP Novel From Placemat To Bowl                      & 60.0 & 57.5 & 44.0 & 74.0 & 54.0 & 56.0 & 62.0 & 76.0 \\
        PnP Novel From Placemat To Plate                     & 56.0 & 63.0 & 48.0 & 70.0 & 64.0 & 80.0 & 52.0 & 58.0 \\
        PnP Novel From Placemat To Tieredshelf               & 36.0 & 28.5 & 18.0 & 26.0 & 38.0 & 14.0 & 24.0 & 32.0 \\
        \midrule
        \rowcolor{gray!20}\textbf{PnP Novel From Placemat To * (Avg)}
                                                              & 45.5 & 51.9 & 38.0 & \underline{54.5} & 46.5 & 48.0 & 48.0 & \textbf{55.5} \\
        \midrule
        PnP Novel From Tray To Cardboardbox                  & 52.0 & 51.5 & 38.0 & 44.0 & 46.0 & 40.0 & 50.0 & 52.0 \\
        PnP Novel From Tray To Plate                         & 48.0 & 71.0 & 56.0 & 66.0 & 72.0 & 66.0 & 58.0 & 68.0 \\
        PnP Novel From Tray To Pot                           & 60.0 & 64.5 & 50.0 & 38.0 & 56.0 & 52.0 & 62.0 & 66.0 \\
        PnP Novel From Tray To Tieredbasket                  & 52.0 & 57.0 & 36.0 & 58.0 & 46.0 & 50.0 & 44.0 & 42.0 \\
        PnP Novel From Tray To Tieredshelf                   & 32.0 & 31.5 & 16.0 & 24.0 & 28.0 & 22.0 & 22.0 & 20.0 \\
        \midrule
        \rowcolor{gray!20}\textbf{PnP Novel From Tray To * (Avg)}
                                                              & 48.8 & \textbf{55.1} & 39.2 & 46.0 & \underline{49.6} & 46.0 & 47.2 & \underline{49.6} \\
        \midrule
        PnP Novel From Plate To Bowl                         & 58.0 & 57.0 & 60.0 & 52.0 & 60.0 & 54.0 & 54.0 & 60.0 \\
        PnP Novel From Plate To Cardboardbox                 & 44.0 & 43.5 & 50.0 & 44.0 & 46.0 & 50.0 & 48.0 & 64.0 \\
        PnP Novel From Plate To Pan                          & 60.0 & 51.0 & 54.0 & 56.0 & 56.0 & 68.0 & 54.0 & 66.0 \\
        PnP Novel From Plate To Plate                        & 64.0 & 78.7 & 70.0 & 62.0 & 66.0 & 78.0 & 78.0 & 76.0 \\
        \midrule
        \rowcolor{gray!20}\textbf{PnP Novel From Plate To * (Avg)}
                                                              & 56.5 & 57.6 & 58.5 & 53.5 & 57.0 & \underline{62.5} & 58.5 & \textbf{66.5} \\
        \midrule
        \rowcolor{gray!30}\textbf{Average}                   & 48.2 & 47.6 & 47.8 & 53.8 & \underline{54.6} & 50.0 & 52.6 & \textbf{57.0} \\
        \bottomrule
    \end{tabular}
    \end{adjustbox}
    \vspace{-0.3em}
\end{table*}

\begin{table*}[!t]
\centering
\footnotesize
\setlength{\tabcolsep}{7pt}
\renewcommand{\arraystretch}{1.18}
\begin{tabularx}{\textwidth}{@{} >{\raggedright\arraybackslash}p{0.22\textwidth} >{\raggedright\arraybackslash}X >{\centering\arraybackslash}p{0.16\textwidth} @{}}
\toprule
\textbf{Family} & \textbf{Task} & \textbf{Max. Test Steps} \\
\midrule
Back-and-forth & Move Block Out and Back & 900 \\
Back-and-forth & Cook Bread and Plate It & 1100 \\
Back-and-forth & Use Stapler and Return It & 900 \\
Back-and-forth & Store Shoe and Take It Back & 900 \\
\midrule
Crossing-path & Move Block to the Opposite Grid & 500 \\
Crossing-path & Move Block to the Other Grid & 500 \\
Crossing-path & Move Phone Between Stand and Pad & 700 \\
\midrule
Bimanual & Hand Over Roller & 900 \\
Bimanual & Hand Over Pill Bottle & 900 \\
\midrule
Multi-goal & Pick Flashed Blocks in Order & 1500 \\
Multi-goal & Pick Flashed Cans in Order & 1200 \\
Multi-goal & Inspect Label and Place Block & 700 \\
\bottomrule
\end{tabularx}
\caption{\textbf{\benchmarkname{} evaluation budgets.} We follow the default RoboTwin2 test protocol and terminate each rollout at task success or at the task-specific maximum number of environment steps. Each maximum test budget is set to twice the corresponding training sampling horizon.}
\label{tab:aliasbench_eval_budget}
\end{table*}

\section{\benchmarkname{} Task Definitions}
\label{sec:appendix_benchmark}

\benchmarkname{} is designed to isolate short-horizon observation aliasing rather than generic long-horizon memory demands. All 12 tasks follow the same core principle: the current frame alone is insufficient to determine the correct continuation, but the missing information is still available in the recent episode context. To isolate this ambiguity, we keep the underlying manipulation difficulty comparable while varying the latent factor that determines the correct continuation, such as the current phase, object source, handoff direction, or a transient cue. We summarize the benchmark families in Table~\ref{tab:aliasbench_overview} and then list the task-level definitions used in our evaluation. 

\paragraph{Evaluation protocol and rollout budgets.}
Following the default RoboTwin2 evaluation protocol~\cite{RoboTwin2_2025_arXiv}, each rollout terminates either when the task succeeds or when it reaches a task-specific maximum number of environment steps. Since the RoboTwin2 report specifies trial-based evaluation but does not prescribe a single global maximum horizon for our newly sampled tasks, we set each maximum test budget to twice the corresponding training sampling horizon. The resulting budgets are listed in Table~\ref{tab:aliasbench_eval_budget}.

\paragraph{Benchmark families.}
The four families are intended to capture common manipulation patterns rather than synthetic edge cases. \textbf{Back-and-forth ambiguity} covers repeated local routines in which nearly identical carrying or staging states reappear in different phases. \textbf{Crossing-path ambiguity} covers source-dependent routing, where similar transport states arise from different recent origins. \textbf{Bimanual ambiguity} captures dual-arm settings in which center or handoff configurations can look nearly symmetric. \textbf{Multi-goal ambiguity} covers scenes with multiple plausible objects or destinations, where a transient cue or recently revealed property selects the active target.

\begin{figure}[!t]
    \centering
    \includegraphics[width=0.98\linewidth]{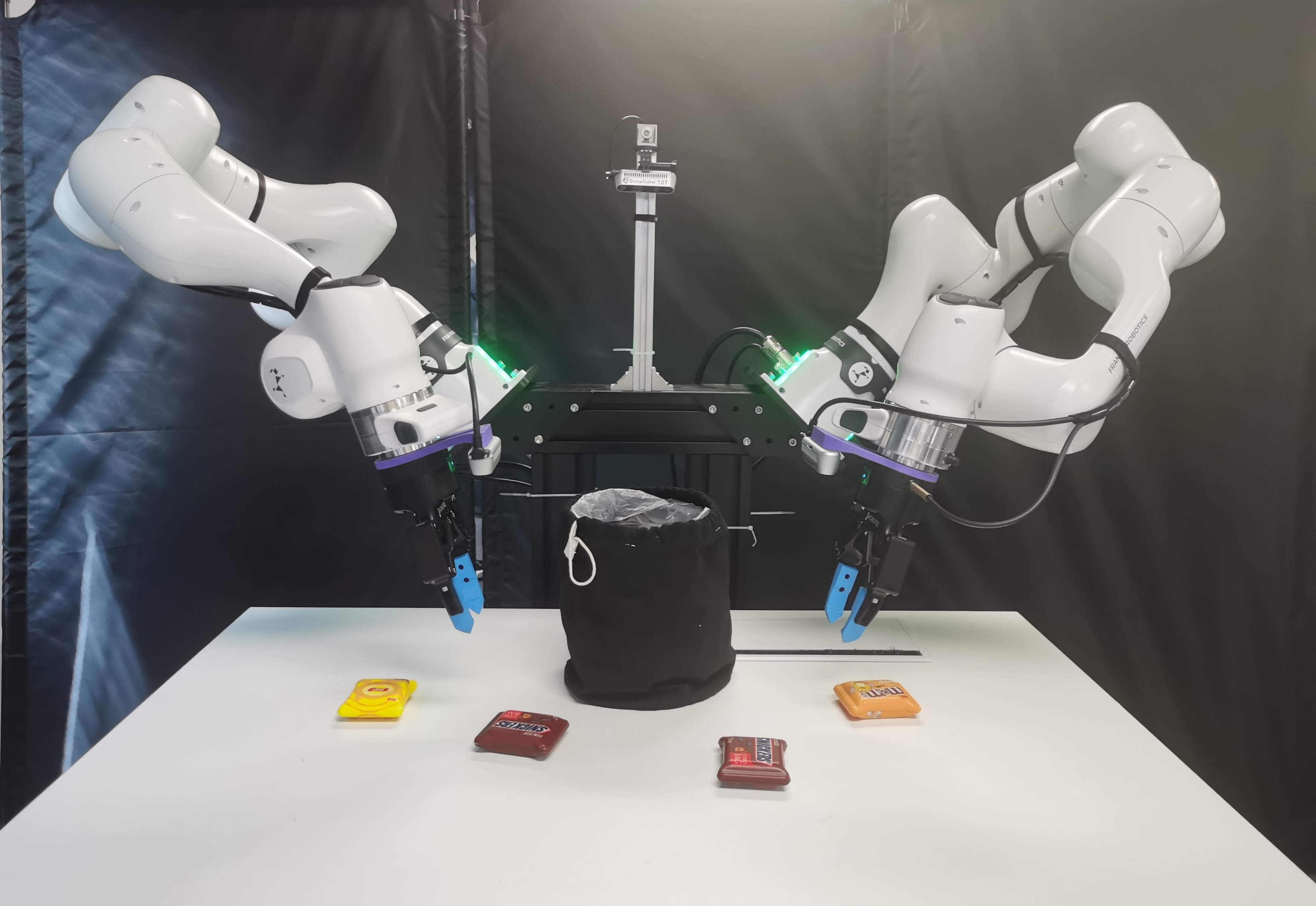}
    \caption{Dual-arm Franka setup for snack cleanup.}
    \label{fig:Franka}
\end{figure}


\paragraph{Representative ambiguity cases.}
Figure~\ref{fig:aliasbench_cases} provides several visual examples from \benchmarkname{}. In \emph{Move Phone Between Stand and Pad}, a natural everyday command is something like ``hey, put the phone on the other stand.'' However, in the third frame, when the robot arm is already holding the phone in mid-air, once one action chunk has finished and the policy must generate the next chunk, the current observation alone no longer reveals which phone stand is the starting point and which one is the target. In \emph{Cook Bread and Plate It}, the first frame, where the robot picks up the bread, and the fifth frame, where it puts the bread down, look similar; likewise, the second frame, where the bread is placed onto the skillet for cooking, and the fourth frame, where it is taken back out and moved for plating, are also visually similar, even though they correspond to different intents. Similarly, in the third frame of \emph{Hand Over Roller}, transferring the roller from left to right and transferring it from right to left can produce similar observations, but the subsequent intents are fundamentally different. These are exactly the kinds of short-horizon aliases that a frame-conditioned chunk policy cannot reliably resolve from the current frame alone. Resolving these aliases therefore requires recent visual context that identifies the active phase, source, or transfer direction.

\begin{table*}[!t]
\centering
\footnotesize
\setlength{\tabcolsep}{7pt}
\renewcommand{\arraystretch}{1.5}
\begin{tabularx}{\textwidth}{@{} >{\raggedright\arraybackslash}p{0.21\textwidth} >{\centering\arraybackslash}p{0.08\textwidth} >{\raggedright\arraybackslash}p{0.20\textwidth} >{\raggedright\arraybackslash}X @{}}
\toprule
Family & \# Tasks & Core latent factor & Definition \\
\midrule
Back-and-forth ambiguity & 4 & Current local phase & Repeated local states recur in different phases of a short routine, but the required next action changes with phase. \\
Crossing-path ambiguity & 3 & Recent source / origin & Similar transport states arise after different pickup origins, and the correct target depends on that recent source. \\
Bimanual ambiguity & 2 & Handoff direction / side of origin & Similar center or handoff states arise in dual-arm transfer, but the next hand and continuation depend on transfer direction. \\
Multi-goal ambiguity & 3 & Transient cue / hidden property & Multiple local candidates coexist, and a brief cue or recently observed property selects the correct target. \\
\bottomrule
\end{tabularx}
\caption{Overview of the four task families in \benchmarkname{}.}
\label{tab:aliasbench_overview}
\end{table*}

\begin{table*}[t]
\centering
\footnotesize
\setlength{\tabcolsep}{7pt}
\renewcommand{\arraystretch}{1.5}
\begin{tabularx}{\textwidth}{@{} >{\raggedright\arraybackslash}p{0.18\textwidth} >{\raggedright\arraybackslash}p{0.25\textwidth} >{\raggedright\arraybackslash}p{0.15\textwidth} >{\raggedright\arraybackslash}X @{}}
\toprule
Task & Task definition & Core latent factor & Why the current frame is insufficient \\
\midrule
\makecell[tl]{Move Block\\Out and Back} & One arm moves a red block from grid A to grid B and then back to A. & Current phase & Similar carrying states appear in both halves, but the next placement differs between the outbound and return phases. \\
\makecell[tl]{Cook Bread\\and Plate It} & A bread slice is moved from a plate to a skillet and then back to the plate. & Current phase & Holding the bread near the workspace can mean placing it on the skillet or returning it to the plate. \\
\makecell[tl]{Use Stapler\\and Return It} & A stapler is moved from pad A to pad B and then back from B to A. & Current phase & Similar transport states recur across two short phases, but the next destination changes with phase. \\
\makecell[tl]{Store Shoe\\and Take It Back} & A shoe is placed into a shoebox and later moved back out to an external target area. & Current phase & Holding the shoe near the box can mean inserting it or taking it back out. \\
\bottomrule
\end{tabularx}
\caption{Back-and-forth ambiguity tasks in \benchmarkname{}.}
\label{tab:aliasbench_backforth}
\end{table*}

\begin{table*}[t]
\centering
\footnotesize
\setlength{\tabcolsep}{7pt}
\renewcommand{\arraystretch}{1.5}
\begin{tabularx}{\textwidth}{@{} >{\raggedright\arraybackslash}p{0.18\textwidth} >{\raggedright\arraybackslash}p{0.25\textwidth} >{\raggedright\arraybackslash}p{0.15\textwidth} >{\raggedright\arraybackslash}X @{}}
\toprule
Task & Task definition & Core latent factor & Why the current frame is insufficient \\
\midrule
\makecell[tl]{Move Block to\\the Other Grid} & A red block starts on one of two grids and must be moved to the other grid. & Recent source & Once the block is in hand, the transport state looks similar for both start grids, but the correct destination is the opposite source grid. \\
\makecell[tl]{Move Block to\\the Opposite Grid} & A red block starts on one of four grids in a 2$\times$2 layout and must be moved to the diagonally opposite grid. & Recent source & Similar mid-transport states occur for different start grids, but the correct diagonal target depends on the pickup origin. \\
\makecell[tl]{Move Phone Between\\Stand and Pad} & A phone begins either on a flat area or on a stand and must be moved to the other location. & Recent source & Carrying the phone looks similar in both cases, while the correct target depends on whether it came from the stand or the flat area. \\
\bottomrule
\end{tabularx}
\caption{Crossing-path ambiguity tasks in \benchmarkname{}.}
\label{tab:aliasbench_crossing}
\end{table*}

\begin{table*}[t]
\centering
\footnotesize
\setlength{\tabcolsep}{7pt}
\renewcommand{\arraystretch}{1.5}
\begin{tabularx}{\textwidth}{@{} >{\raggedright\arraybackslash}p{0.18\textwidth} >{\raggedright\arraybackslash}p{0.25\textwidth} >{\raggedright\arraybackslash}p{0.15\textwidth} >{\raggedright\arraybackslash}X @{}}
\toprule
Task & Task definition & Core latent factor & Why the current frame is insufficient \\
\midrule
\makecell[tl]{Hand Over\\Pill Bottle} & A pill bottle starts on the left or right outer pad, moves to the center pad, and is then sent to the opposite outer pad by the other hand. & Handoff continuation & At the center pad, both hands and the bottle can look similar, but the correct next hand depends on which side the bottle came from. \\
\makecell[tl]{Hand Over\\Roller} & A long roller starts on one side, is moved to the centerline, handed over, and then placed on the opposite side. & Handoff direction & Near the centerline, the handoff state is nearly symmetric, yet the continuation depends on transfer direction. \\
\bottomrule
\end{tabularx}
\caption{Bimanual ambiguity tasks in \benchmarkname{}.}
\label{tab:aliasbench_bimanual}
\end{table*}

\begin{table*}[t]
\centering
\footnotesize
\setlength{\tabcolsep}{7pt}
\renewcommand{\arraystretch}{1.5}
\begin{tabularx}{\textwidth}{@{} >{\raggedright\arraybackslash}p{0.18\textwidth} >{\raggedright\arraybackslash}p{0.25\textwidth} >{\raggedright\arraybackslash}p{0.15\textwidth} >{\raggedright\arraybackslash}X @{}}
\toprule
Task & Task definition & Core latent factor & Why the current frame is insufficient \\
\midrule
\makecell[tl]{Pick Flashed\\Blocks in Order} & Three blocks are available, and the object to be picked briefly flashes underneath before being placed onto the current target pad. & Active-target cue & After the flash disappears, several candidate blocks remain, but the frame no longer reveals which one is active. \\
\makecell[tl]{Pick Flashed\\Cans in Order} & Two cans are presented, and the can to be picked briefly flashes underneath before being placed onto the current target pad. & Active-target cue & Once the flash vanishes, multiple candidate picks remain plausible in the current frame. \\
\makecell[tl]{Inspect Label\\and Place Block} & The robot inspects a hidden label on a block and then places it onto the matching target area instead of a distractor area. & Hidden property & During final transport and placement, the label may no longer be visible, so the target depends on the recent inspection result. \\
\bottomrule
\end{tabularx}
\caption{Multi-goal ambiguity tasks in \benchmarkname{}.}
\label{tab:aliasbench_multigoal}
\end{table*}

\section{Real-world robot setup.}
\begin{figure*}[b]
    \centering
    \includegraphics[width=1\linewidth]{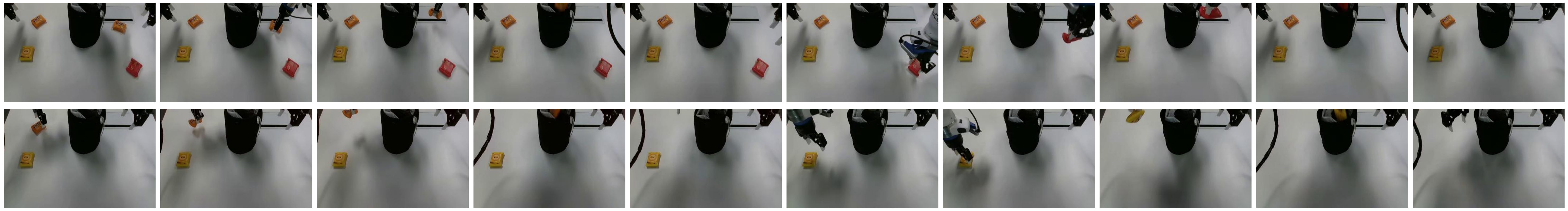}
    \caption{\textbf{Representative trajectory from the real-world snack-cleanup task.} The sequence demonstrates \methodname{} executing a multi-snack collection episode. This trajectory represents an upper bound of our model's capabilities.}
    \label{fig:real_world_snack_demo}
\end{figure*}

Figure~\ref{fig:Franka} shows the real-world snack-cleanup setup: two Franka Research 3 arms with Robotiq 85 grippers. Intel RealSense D435 cameras capture observations from head and end-effector views. We use Polymetis~\cite{polymetis}~for teleoperation and collect human expert demonstrations.
We also present a representative trajectory in Figure~\ref{fig:real_world_snack_demo}.

\end{document}